\documentclass[a4paper,fleqn]{cas-sc}

\usepackage[authoryear]{natbib}
\usepackage{graphicx}%
\usepackage{multirow}%
\usepackage{amsmath,amssymb,amsfonts}%
\usepackage{tabularx} %
\usepackage{booktabs} %
\usepackage{multirow} %
\usepackage{threeparttable}
\usepackage{amsthm}%
\usepackage{mathrsfs}%
\usepackage{xcolor}%
\usepackage{textcomp}%
\usepackage{manyfoot}%
\usepackage{booktabs}%
\usepackage{algorithm}%
\usepackage{algorithmicx}%
\usepackage{algpseudocode}%
\usepackage{listings}%
\usepackage{silence}
\usepackage{url}
\usepackage{booktabs}
\usepackage{amssymb}
\usepackage{cite}
\usepackage{amsmath,amssymb,amsfonts}
\usepackage{graphicx}
\usepackage{textcomp}
\usepackage{algorithm}
\usepackage{algpseudocode}
\usepackage{listings} 
\usepackage{color} 
\usepackage{booktabs}
\usepackage{amsmath}
\usepackage{multirow}
\usepackage{listings}
\usepackage{algorithm}
\usepackage{booktabs}
\def\tsc#1{\csdef{#1}{\textsc{\lowercase{#1}}\xspace}}
\tsc{WGM}
\tsc{QE}
\tsc{EP}
\tsc{PMS}
\tsc{BEC}
\tsc{DE}

\begin{document}
\let\WriteBookmarks\relax
\def\floatpagepagefraction{1}
\def\textpagefraction{.001}

\shorttitle{mChartQA}

\shortauthors{Jingxuan Wei et~al.}

\title [mode = title]{mChartQA: A universal benchmark for multimodal Chart Question Answer based on Vision-Language Alignment and Reasoning}                      

\author[inst1,inst3]{Jingxuan Wei}
\ead{weijingxuan20@mails.ucas.edu.cn}

\cortext[corcorr]{Corresponding author.}
\author[inst2]{Nan Xu}
\ead{nan.xu@wenge.com}
\cormark[1]

\author[inst1,inst3]{Guiyong Chang}
\ead{changguiyong22@mails.ucas.ac.cn}

\author[inst2]{Yin Luo}
\ead{yin.luo@wenge.com}

\author[inst1,inst3]{Bihui Yu}
\ead{yubihui@sict.ac.cn}

\author[inst1,inst3]{Ruifeng Guo}
\ead{guofuifeng@163.com}

\affiliation[inst1]{organization={Shenyang Institute of Computing Technology, Chinese Academy of Sciences},
    city={Shenyang},
    postcode={110168}, 
    country={China}}

\affiliation[inst2]{organization={Beijing Wenge Technology Co., Ltd},
    city={Beijing},
    postcode={100049}, 
    country={China}}

\affiliation[inst3]{organization={University of Chinese Academy of Sciences},
    city={Beijing},
    postcode={100049}, 
    country={China}}

\begin{abstract}
Multimodal chart question-answering, crucial for applications such as financial report analysis, decision support, and invoice parsing, confronts significant challenges with intricate color patterns, structural complexities, and implicit numerical data in charts. Traditional methods, mainly involving chart-to-text conversion followed by processing with Large Language Models (LLMs) or direct multimodal processing, often falter in these complex scenarios. To overcome these hurdles, this paper introduces mChartQA, a groundbreaking framework tailored for advanced multimodal chart question-answering. mChartQA innovatively merges sophisticated language processing capabilities of LLMs with a state-of-the-art table-to-text engine, facilitating effective processing and integration of complex visual and textual information. This framework stands out for its ability to align visual and textual data accurately and is further refined for deep reasoning and contextual understanding within charts. The AI contribution of this work lies in its novel integration of multimodal data processing techniques, significantly enhancing the accuracy of chart question-answering. Demonstrated through experimental results on three distinct datasets, mChartQA showcases superior performance in tackling complex, multimodal chart question-answering tasks, especially in scenarios that have posed challenges for existing methods.
\end{abstract}

\begin{keywords}
Multimodal Chart Question-Answering\sep Vision-Language Alignment\sep Natural Language Processing\sep Large Language Models (LLMs)\sep Two-Stage Training
\end{keywords}

\maketitle

\section{Introduction}
The goal of multimodal chart question answering is to automatically answer a natural language question about a chart to facilitate visual data analysis ~\citep{cqa}, where the ability to understand and interact with visual data is essential ~\citep{chartqa}. It has emerged as a crucial intersection of computer vision and natural language processing, addressing the growing demand for intelligent systems capable of interpreting complex visual data in charts ~\citep{chartqa}. Beyond its general applications, multimodal chart question-answering plays a pivotal role in sectors requiring precise and rapid analysis of visual data. In the financial domain, it is indispensable for tasks such as financial report analysis ~\citep{wang2023attentive}, decision support~\citep{datadriven}, invoice parsing ~\citep{gerling2023multimodal}, and contract review~\citep{jie2023novel}. Similarly, in the medical field, it significantly contributes to the digitization of patient records~\citep{xu2021mufasa}, medical insurance review~\citep{mesko2023impact}, diagnostic assistance~\citep{othmani2022multimodal}, and quality control~\citep{schilcher2024fusion} of medical records.

\begin{figure}[t]
\centering
\includegraphics[width=0.6\linewidth]{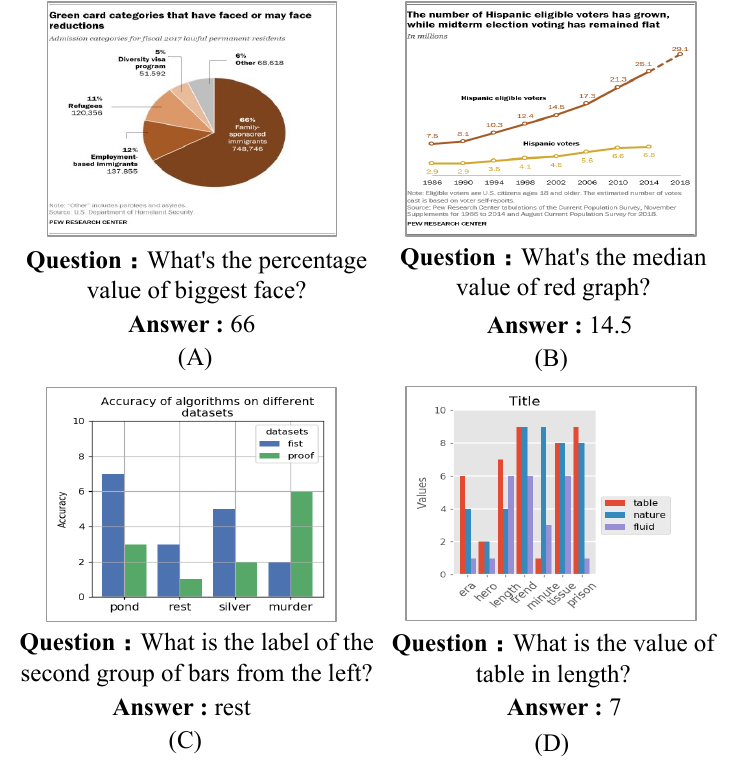}
\caption{Examples of Color, Structure, and Textless Charts}
\label{fig:example_charts}
\end{figure}

Due to the richness and ambiguities of natural language and complex visual reasoning, multimodal chart question answering task requires to predict the answer in the intersection of information visualization, natural language processing, and human computer interactions ~\citep{cqa}. 

Early approaches apply natural language processing techniques by largely depending on heuristics or grammar-based parsing techniques ~\citep{Eviza,Orko,Applying,DataTone}. Thanks to insufficient processing of complex linguistic phenomena, over-reliance on grammatical rules, and limited depth of understanding natural language,  deep learning models have been introduced for understanding natural language queries about visualizations ~\citep{LEAF-QA,STL-CQA,FigureNet}.

Recently, with the outstanding performance of large language models (LLMs) in natural language inference, researches propose a pipeline approach which involve converting chart information into textual format, known as chart-to-text, and then processing this information using LLMs ~\citep{deplot}. Though it performs well in text-based scenarios (seen Fig.\ref{fig:example_charts}(A)), this two-stage approach tends to struggle when it comes to complex scenarios such as colors, structures, or textless information (seen Fig.\ref{fig:example_charts}(B)(C)(D)). The main challenge here is that crucial visual details may be lost or misrepresented during the chart-to-text conversion process, leading to potential misunderstandings or errors in interpretation. Hence, researches further explore the multimodal alignment framework based on  pre-trained vision-language (VL) models that directly process the visual form of charts, such as mPLUG-DocOwl~\citep{mplug-docowl} and Qwen-VL ~\citep{qwenvl}. Despite their strong alignment capabilities in VL tasks, these models reveal limitations in complex reasoning scenarios within chart question answering, particularly in handling questions about color patterns in charts, structural details, and numerical data, especially in instances where charts do not explicitly display numerical information.

Given the limitations of existing methods in multimodal chart question answering, particularly in scenarios that require intricate understanding of color patterns, structural complexities, and interpretation of charts with implicit numerical data, our research is focused on developing a more adaptive and comprehensive solution. This need underscores the importance of a model capable of effectively processing and integrating diverse visual and textual information. 
In response, we propose a multimodal chart question answering framework, mChartQA, which leverages vision-language alignment and reasoning. This framework distinctively integrates advanced language processing techniques from large language models (LLMs) with a sophisticated table-to-text engine, enabling the transformation of complex visual elements into analytically rich formats. Through this integration, mChartQA aims to overcome the limitations of current VL models, enhancing their ability to provide accurate and contextually rich answers to complex questions about multimodal charts.
The main contributions of this paper are as follows:
\begin{itemize}
    \item We propose a universal benchmark for multimodal chart question answer based on vision-language alignment and reasoning, which provides the reasoning ability and word-level interpretability for multimodal chart question answer task.
    \item Our proposed mChartQA aligns visual and textual data to ensure accurate interpretation of visual elements, and then applies advanced language processing techniques for deep reasoning and contextual understanding.
    \item We compare mChartQA with existing state-of-the-art methods, and experimental results on three datasets demonstrate effectiveness of our model in handling diverse and challenging chart question-answering tasks.
\end{itemize}

\section{Related Work}

The field of chart question-answering has evolved significantly, beginning with early reliance on natural language processing (NLP) techniques and advancing towards sophisticated multimodal approaches. Initially, chart question-answering systems like Eviza ~\citep{Eviza}, Orko ~\citep{Orko}, Evizeon ~\citep{Applying} and DataTone ~\citep{DataTone} employed heuristic or grammar-based parsing techniques. While these methods provided foundational insights, they struggled with complex linguistic phenomena and heavily relied on grammatical rules.
To address these limitations, more advanced models such as LEAF-QA ~\citep{LEAF-QA}, STL-CQA ~\citep{STL-CQA}, and FigureNet ~\citep{FigureNet} were developed. LEAF-QA introduced a dataset of figures/charts with question-answer pairs for figure question answering, constructed from real-world open data sources ~\citep{LEAF-QA}. STL-CQA and FigureNet, on the other hand, focused on deep learning models for question-answering on scientific plots, pushing the boundaries in reasoning and understanding tasks ~\citep{STL-CQA, FigureNet}.

The emergence of benchmarks like FigureQA ~\citep{figureqa}, PlotQA ~\citep{plotqa}, and ChartQA ~\citep{chartqa} further highlighted the need for advanced multimodal methods in chart interpretation. These benchmarks have been instrumental in driving research towards more integrated approaches that combine linguistic and visual analysis. In response to these challenges, two main approaches have emerged in the field of chart question-answering:
\textit{1)Chart-to-Text Conversion and LLM Processing: }
The chart-to-text conversion approach, exemplified by DePlot ~\citep{deplot}, involves translating visual chart data into text for processing with large language models (LLMs). This method is effective for simpler visual scenarios but often struggles with the complexity of visual elements in more intricate charts.
\textit{2)Direct Multimodal Processing:}
In contrast, direct multimodal processing using vision-language (VL) models has gained prominence. Models like Pix2Struct ~\citep{pix2struct}, PaLI-3 ~\citep{pali3}, BLIP-2 ~\citep{blip2}, Qwen-VL ~\citep{qwenvl}, mPLUG-DocOwl ~\citep{mplug-docowl}, and UniChart ~\citep{unichart} have been at the forefront of this research, exploring innovative architectures for chart comprehension and reasoning. 
However, despite their strong  capabilities in VL tasks, these models often encounter limitations in complex reasoning scenarios within chart question answering.

\section{Method}

\begin{figure*}
    \centering
    \includegraphics[width=\linewidth]{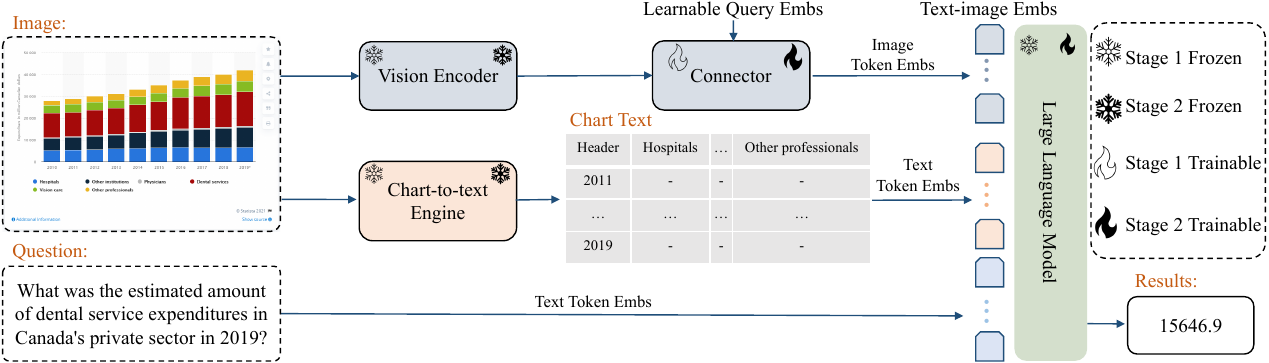}
    \caption{The training architecture and workflow of the mChartQA model.}
    \label{fig:new_model_architecture}
\end{figure*}
\subsection{Architecture}
Our model, mChartQA (Multimodal Chart Question-Answering Model), is designed for aligning and reasoning with visual and textual information from chart images and corresponding questions. The architecture, illustrated in Figure \ref{fig:new_model_architecture}, includes four main components:

\textbf{Vision Encoder (E\textsubscript{v}):} 
The Vision Encoder processes a chart image \(I\) to produce visual features \(V\). This process is formalized as \(V = E_v(I)\), capturing detailed visual information from the chart image.

\textbf{Connector (C):} 
The Connector employs a cross-attention mechanism to align visual features \(V\) with the text encoder. The alignment process is crucial for correlating visual elements with corresponding textual data, enhancing the model's interpretative capabilities. The cross-attention mechanism in the Connector is defined as:
\begin{equation}
\scalebox{0.9}{
    $V' = C(V) = \text{softmax}\left(\frac{(W_qV)(W_kV)^T}{\sqrt{d_k}}\right)(W_vV),$
}
\end{equation}
where \(Q = W_qV\) is a learnable query, \(K = W_kV\) and \(V = W_vV\) are key and value projections with \(W_k\) and \(W_v\) being the respective weights, and \(d_k\) is the dimension of the key vectors.

\textbf{Chart-to-Text Engine (T):} 
This module converts the chart image \(I\) into a textual representation \(T'\), formalized as \(T' = T(I)\), extracting key textual elements from the chart. 
For instance, in Figure \ref{fig:new_model_architecture}, if the bar chart lacks numerical annotations, the engine will only recognize and extract the definite textual information present, without performing predictions based on the bar chart.

\textbf{Large Language Model (L):} 
The Large Language Model processes the tokenized question \(Q_t\), enhanced visual features \(V'\), and tokenized textual representation \(T'_t\). The prediction process is formalized as \(A = L(Q_t, V', T'_t)\), where \(L\) integrates visual and textual information to predict the answer.

\subsection{Training}
mChartQA training is conducted in two stages:

\textbf{Stage 1 - Visual-Language Alignment:}
This stage focuses on training the Connector to optimize the alignment of visual and textual representations. The objective function for this stage is defined as:
\begin{equation}
\scalebox{0.9}{
    $\min_{\theta_C} \mathcal{L}_{\text{alignment}}(V, Q_t; \theta_C) = -\sum_{i=1}^{N} \log P(y_i | V, Q_t; \theta_C),$
}
\end{equation}
where \(\mathcal{L}_{\text{alignment}}\) is the cross-entropy loss, \(y_i\) are the true labels, and \(P(y_i | V, Q_t; \theta_C)\) is the predicted probability of the correct label.

\textbf{Stage 2 - Visual-Language Reasoning:}
In this stage, both the Connector and the Large Language Model are trained to enhance reasoning capabilities. The final optimization objective is defined as:
\begin{equation}
\scalebox{0.75}{
    $\min_{\theta_C, \theta_L} \mathcal{L}_{\text{reasoning}}(Q_t, V', T'_t; \theta_C, \theta_L) = -\sum_{i=1}^{N} \log P(y_i | Q_t, V', T'_t; \theta_C, \theta_L)$
}
\end{equation}
where \(\mathcal{L}_{\text{reasoning}}\) is the cross-entropy loss, \(y_i\) are the true labels, and \(P(y_i | Q_t, V', T'_t; \theta_C, \theta_L)\) is the predicted probability of the correct label.

\section{Experiment}
\subsection{Datasets} 
\label{dataset}

\begin{table}[htbp]
  \centering
  \caption{Details of stage one training data}
  \resizebox{0.6\textwidth}{!}{
    \begin{tabular}{clc}
    \toprule
    \textbf{Task} & \textbf{Dataset} & \textbf{Used}  \\
    \midrule
                &COCO Caption~\citep{cococaption}  &400K \\
                &SBU~\citep{sbu} &  300K  \\
   Captioning   & NoCaps~\citep{nocaps}& 200K\\
                & CC3M~\citep{cc3m} & 200K\\
                & ShareGPT4V~\citep{sharegpt4v}& 500K  \\
    \midrule
                & GRIT~\citep{GRIT} &150K \\
                & Visual Genome~\citep{VisualGenome} & 100K \\
   Grounding   & RefCOCO~\citep{RefCOCO}& 50K\\
                & RefCOCO+~\citep{RefCOCO+} & 50K\\
                & RefCOCOg~\citep{RefCOCO+}&50K  \\
    \midrule
    Chart-to-text & ChartQA ~\citep{chartqa}&  20K \\
    \midrule
             & Total&  2.02M \\
    \bottomrule
    \end{tabular} }
  \label{tab:Stage1Data}%
\end{table}%

\begin{figure}[!t]\centering
        \centering
	\includegraphics[width=0.8\linewidth]{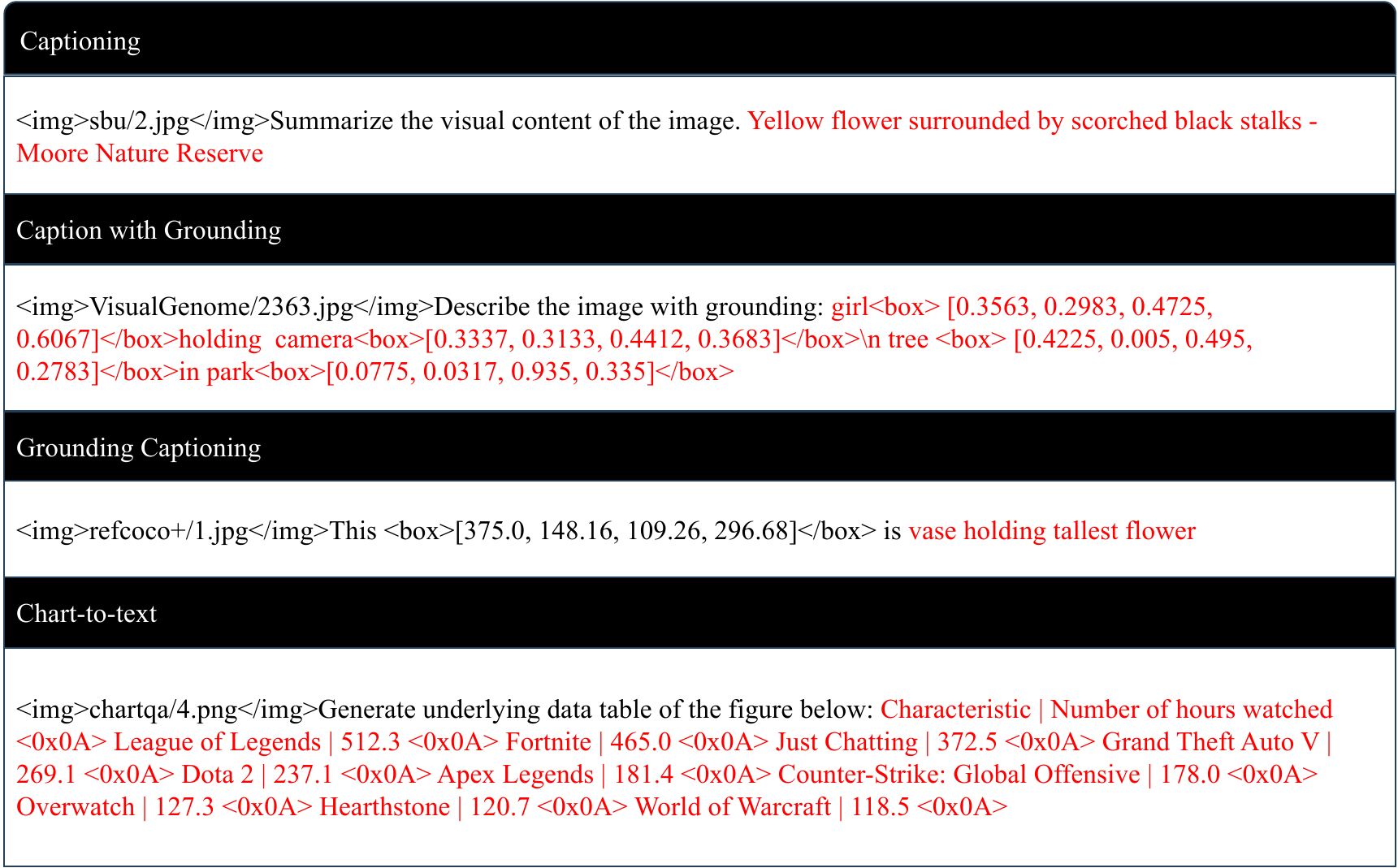}
	\caption{Format for Stage 1 training data includes Captioning, Grounding, and Chart-to-Text tasks. The prefix sequence is in black text, while the correct label is in red text.}
        \label{fig:stage1_dataset}
\end{figure}

\textit{Stage 1 - Visual-Language Alignment}
In the initial stage, we mainly focus on realizing Image-text alignment and improving its fine-grained perception of images. Our model was trained using data specific to the Captioning, Grounding, and Chart-to-text tasks, as illustrated in Table \ref{tab:Stage1Data}. 
We primarily utilize 1600,000 text pairs for training purposes in the captioning task. For the grounding task, we utilized 400,000 data points for training and bifurcated it into two subtasks, Caption with Grounding and Grounded Captioning. In the Caption with Grounding task, the model is required to accurately identify and describe a specific object in the image while simultaneously labeling the position (box) of the object in the image. On the other hand, in the Grounded Captioning task, the model must describe the object according to the information provided by its position (box) in the image. These two tasks mutually benefit one another, greatly enhancing the model's capability to detail objects within images at an intricate level. Additionally, We incorporated information from the 20,882 charts provided by the chartqa dataset to facilitate training on the chart-to-text task. Figure \ref{fig:stage1_dataset} demonstrates the precise structure of these aforementioned tasks.
Through implementation of the above practices, our model can attain exceptional chart-to-text alignment and exhibit a profound comprehension of charts. This training methodology serves to notably enhance the model's faculty for discerning images at a precise level, thereby yielding more exact and comprehensive information for ensuing analyses and processing endeavors.

\textit{Stage 2 - Visual-Language Reasoning:}  In the stage, we utilized methods similar to ~\citep{xia2024chartx,meng2024chartassisstant}. We randomly extracted data from several training datasets for our study: 10,000 pairs from the ChartQA training dataset, 140,000 pairs from the FigureQA training dataset, and 150,000 pairs from the PlotQA training dataset, totaling 300,000 pairs.
The table \ref{tab:dataset_summary} summarizes the datasets used and the number of chart-question pairs originally available versus the number randomly extracted for our study.

\begin{table}[H]
\centering
\begin{tabular}{lcc}
\hline
\textbf{Dataset} & \textbf{Original Number of Chart-Question Pairs} & \textbf{Number of Pairs Randomly Extracted} \\
\hline
ChartQA & 28,299 & 10,000 \\
FigureQA & 2,388,698 & 140,000 \\
PlotQA & 28,952,641 & 150,000 \\
\hline
\end{tabular}
\caption{Summary of datasets and the number of chart-question pairs extracted.}
\label{tab:dataset_summary}
\end{table}

\textit{Test DataSets:} We conduct tests on the public test sets of three datasets: ChartQA, PlotQA, and FigureQA. Similar to the approach in ~\citep{wang2023domino}, our test set composition is detailed below. We selected a set of representative problems from these datasets to evaluate the model's performance on color, structure, and textless problems in more detail. These datasets present unique challenges, including the complexity of color patterns, structures, and the graphical interpretation of implicit numerical data. This method of evaluation provides a more detailed understanding of model performance. We identify three main types of questions, with their types and example templates described below.

\textbf{Color:} This problem type requires an understanding of color theory, as well as the observation and analysis of color information on a chart. Example: "What is the least difference between the light blue bar and the dark blue bar?"

\textbf{Structure:} These problems relate to chart layout and structure, requiring an analysis and understanding of the components of a chart and its visual representation. Example: "What is the label of the third bar from the left in each group?"

\begin{figure*}[htbp]\centering
        \centering
	\includegraphics[width=\linewidth]{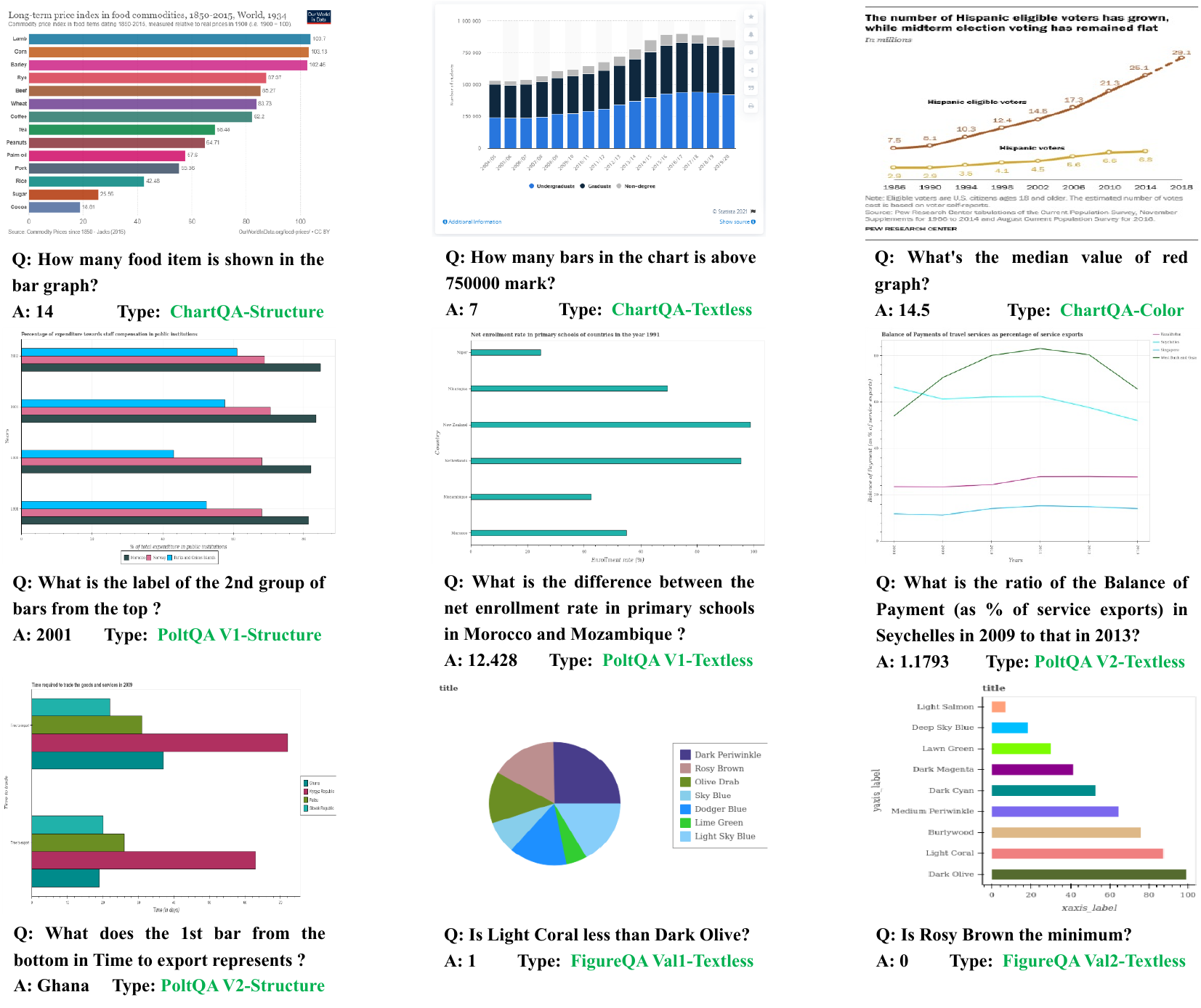}
	\caption{Example test dataset extracted from the ChartQA, PlotQA, and FigureQA datasets, with the test set and example type displayed in green.}
        \label{fig:test_dataset2}
\end{figure*}

\textbf{Textless:} This problem involves interpreting graphs containing implicit numerical data, where the graphical elements lack precise numerical values and require numerical reasoning using a model. To enhance the construction of the test dataset, we manually filtered out datasets containing these three types of queries from the ChartQA, PlotQA, and FigureQA test dataset. These samples are visible in Figure \ref{fig:test_dataset2}. The diversity of the test dataset demonstrates the model's ability to address diagrammatic problems, especially those involving color, structure, and textless diagrams, serving as an important evaluation criterion.

\begin{table}[htbp]
  \centering
  \caption{Comprehensive Distribution of QA Pairs and Problem Types Across Test Datasets}
  \resizebox{0.6\textwidth}{!}{
    \begin{tabular}{l|c|ccc}
    \toprule
    \textbf{Dataset} & \textbf{QA Pairs} & \textbf{Color} & \textbf{Structure} & \textbf{Textless} \\
    \midrule
    ChartQA & 2,500 & 264 & 385 & 209 \\
    PlotQA V1 & 10,000 & --- & 966 & 1,165 \\
    PlotQA V2 & 10,000 & --- & 310 & 662 \\
    FigureQA Val1 & 5,000 & --- & --- & 1,607 \\
    FigureQA Val2 & 5,000 & --- & --- & 1,627 \\
    \bottomrule
    \end{tabular}
  }
  \label{tab:comprehensive_distribution}
\end{table}

\subsection{Baselines}
In our experiments with mChartQA, we compare it against two categories of baseline models:
\textit{Few-Shot Learning Models:} We include pioneering models such as GPT3 (1-Shot) ~\citep{wang2023domino} and GPT4 (5-Shot) ~\citep{wang2023domino}, which leverage generative pre-trained transformers for Few-Shot learning. Additionally, FlanPaLM (540B) in both 1-Shot ~\citep{deplot} and 1-Shot with Self-Consistency (SC) decoding ~\citep{deplot}, as well as LLaMA-2 (70B) in 1-Shot and 5-Shot configurations ~\citep{wang2023domino}, are evaluated to highlight their Few-Shot capabilities in the context of chart question-answering.
\textit{Fully-Supervised Models:} This category includes specialized models such as Pix2Struct ~\citep{pix2struct} and MATCHA ~\citep{matcha}, which are designed specifically for interpreting and answering questions from charts. Large Language Model (LLM)-Based Approaches like PaLI-X-OCR ~\citep{chen2023pali} and PaLI-X-no\_OCR ~\citep{chen2023pali}, mPLUG-DocOwl (LLaMA-7B) ~\citep{ye2023mplug}, Qwen-VL~\citep{qwenvl}, and Qwen-VL-Chat ~\citep{qwenvl} demonstrate the adaptability of pre-trained models to chart question-answering. Vision-Language Pretraining (VLP) models such as VL-T5-OCR ~\citep{deplot}, T5-OCR ~\citep{deplot}, and VisionTapas-OCR ~\citep{deplot}, along with ChartReader ~\citep{cheng2023chartreader} and ChartT5 ~\citep{zhou2023enhanced}, showcase the integration of OCR capabilities for enhanced performance. Additionally, the DOMINO (70B) ~\citep{wang2023domino} is included to represent a state-of-the-art approach in fully supervised learning for chart question-answering.
Through this array of baselines, we aim to provide a comprehensive evaluation of mChartQA's performance, situating it within the broader landscape of chart question-answering research.


\subsection{Experimental Setting}
In our experiments, we have developed two main versions of the mChartQA model: mChartQA$_{\text{Qwen}}$ and mChartQA$_{\text{Intern-LM2}}$. The mChartQA$_{\text{Qwen}}$ version is based on the Qwen-14B Large Language Model, while the mChartQA$_{\text{Intern-LM2}}$ version utilizes the Intern-LM2-7B model. Both versions employ the clip-vit-large-patch14-336 as their vision encoder, ensuring advanced visual processing capabilities. The Chart-to-Text Engine across both versions is initialized with weights from DePlot, facilitating effective translation of chart images into descriptive text.

\textbf{Training Configuration:}
\textit{1) Visual-Language Alignment } 
In this stage, we use the AdamW optimizer with \(\beta_1 = 0.9\), \(\beta_2 = 0.95\), a learning rate of \(1e-6\), weight decay of \(0.001\), and a batch size of 16. The training spans across 6 epochs, focusing on aligning visual and textual representations. 
\textit{2) Visual-Language Reasoning } 
In the reasoning stage, we continue with the same optimizer and weight decay settings but adjust the learning rate to \(1e-5\) and reduce the batch size to 8. This stage lasts for 8 epochs and involves training both the Connector and the Large Language Model to enhance reasoning capabilities.

\textbf{Evaluation Metrics:}
Our evaluation follows the metrics used in the ChartQA, PlotQA, and FigureQA datasets. For numeric answers, a tolerance of up to 5\% error is allowed. However, for textual answers, exact matching is required. Only responses that precisely match the correct answer are considered correct, ensuring accuracy in evaluation.

\subsection{Main Results}

\begin{table}[htbp]
   \centering
    \caption{Extended Comparative Results on the ChartQA Dataset: The columns represent the model name, performance on human-generated questions (ChartQA-human), performance on machine-generated questions (ChartQA-M), and the average performance across both types.}
   {\renewcommand\baselinestretch{1.25}\selectfont
    \resizebox{\textwidth}{!}{
    \begin{tabular}{lcccc}
     \toprule
    \textbf{Evaluation Type} & \textbf{Model} & \textbf{ChartQA-human (1250)} & \textbf{ChartQA-M (1250)} & \textbf{AVG} \\
      \midrule
    \multirow{6}{*}{Few-Shot} & GPT3 (1-Shot) ~\citep{deplot} & 36.50 & 37.30 & 36.90 \\
     & FlanPaLM (540B) (1-Shot)~\citep{deplot} & 57.80 & 76.70 & 67.30 \\
     & FlanPaLM (540B) (1-Shot, SC)~\citep{deplot} & 62.20 & 78.80 & 70.50 \\
     & LLaMA-2 (70B) ~\citep{wang2023domino} (1-Shot) & 53.50 & 86.50 & 70.00 \\
     & GPT4 (5-Shot) ~\citep{wang2023domino} & 61.40 & 83.80 & 72.60 \\
     & LLaMA-2 (70B) (5-Shot) ~\citep{wang2023domino} & 59.40 & 87.40 & 73.40 \\
      \midrule
    \multirow{15}{*}{Fully-Supervised} 
     & PaLI-X-OCR ~\citep{chen2023pali} & - & - & 72.30 \\
     & PaLI-X-no\_OCR~\citep{chen2023pali} & - & - & 70.90 \\
     & mPLUG-DocOwl (LLaMA-7B) ~\citep{ye2023mplug} & - & - & 57.40 \\
     & Qwen-VL (Qwen-7B) ~\citep{qwenvl} & - & - & 65.70 \\
     & Qwen-VL-Chat ~\citep{qwenvl} & - & - & 66.30 \\
     & VL-T5-OCR ~\citep{deplot} & - & - & 41.60 \\
     & T5-OCR ~\citep{deplot} & - & - & 41.00 \\
     & VisionTapas-OCR ~\citep{deplot}  & - & - & 45.50 \\
     & ChartReader ~\citep{cheng2023chartreader} & - & - & 52.60 \\
     & ChartT5 ~\citep{zhou2023enhanced} & 31.80 & 74.40 & 53.20 \\
     & Pix2Struct~\citep{pix2struct} & 30.50 & 81.60 & 56.10 \\
     & MATCHA ~\citep{matcha} & 38.20 & 90.20 & 64.20 \\
     & PaLI-17B ~\citep{chen2023pali} (res. 588) & 30.40 & 64.90 & 47.60 \\
     &UniChart~\citep{unichart}& 43.92 & 88.56 & 66.24\\
     & DOMINO(70B) ~\citep{wang2023domino} & 64.10 & 91.80 & 78.00 \\
     \midrule
     \multirow{2}{*}{Ours} & \cellcolor[HTML]{DEDFDF}mChartQA$_{\text{Qwen}}$ & \cellcolor[HTML]{DEDFDF}\textbf{58.56} & \cellcolor[HTML]{DEDFDF}\textbf{93.44} & \cellcolor[HTML]{DEDFDF}\textbf{76.00} \\
& \cellcolor[HTML]{DEDFDF}mChartQA$_{\text{Intern-LM2}}$ & \cellcolor[HTML]{DEDFDF}\textbf{68.24} & \cellcolor[HTML]{DEDFDF}\textbf{89.76} & \cellcolor[HTML]{DEDFDF}\textbf{79.00} \\
    \bottomrule
    \end{tabular}} \par}
 \label{tab:extended_chartqa}
\end{table}%

\begin{table}[htbp]
   \centering
   \caption{Model Performance on FigureQA and PlotQA Datasets: Accuracy scores for Few-Shot and Supervised models on FigureQA v1, v2, and their average (FigureQA Avg), followed by PlotQA v1, v2, and their average (PlotQA Avg).}
   \resizebox{\textwidth}{!}{
    \begin{tabular}{lccccccc}
     \toprule
    \textbf{Evaluation Type} & \textbf{Model} & \textbf{FigureQA v1} & \textbf{FigureQA v2} & \textbf{FigureQA Avg} & \textbf{PlotQA v1} & \textbf{PlotQA v2} & \textbf{PlotQA Avg} \\
      \midrule
    \multirow{5}{*}{Few-Shot} & GPT3 (1-Shot) ~\citep{deplot} & - & - & - & 31.60 & 42.20 & 36.90 \\
     & FlanPaLM (540B) (1-Shot) ~\citep{deplot} & - & - & - & 51.30 & 44.90 & 48.10 \\
     & FlanPaLM (540B) (1-Shot, SC) ~\citep{deplot} & - & - & - & 57.80 & 50.10 & 53.90 \\
     & LLaMA-2 (70B) (1-Shot) ~\citep{wang2023domino} & 55.60 & 55.70 & 55.65 & 32.50 & 43.40 & 37.90 \\
     & LLaMA-2 (70B) (5-Shot) ~\citep{wang2023domino} & 61.60 & 61.20 & 61.40 & 43.20 & 44.70 & 43.90 \\
      \midrule
    \multirow{3}{*}{Supervised} & Pix2Struct ~\citep{pix2struct} & - & - & - & 73.20 & 71.90 & 72.60 \\
     & MATCHA  ~\citep{matcha}& 50.02 & 50.10 & 50.06 & 77.53 & 60.30 & 68.92 \\
     & ChartReader ~\citep{cheng2023chartreader} & 95.50 & 95.80 & 95.65 & 78.10 & 59.30 & 68.70 \\
     & DOMINO(70B) ~\citep{wang2023domino} & 64.70 & 64.40 & 64.55 & 58.90 &  80.70 & 69.80 \\
     \midrule
     \multirow{2}{*}{Ours}
     &\cellcolor[HTML]{DEDFDF}mChartQA$_{\text{Qwen}}$ & \cellcolor[HTML]{DEDFDF}\textbf{90.32} & \cellcolor[HTML]{DEDFDF}\textbf{92.75} & \cellcolor[HTML]{DEDFDF}\textbf{91.54} & \cellcolor[HTML]{DEDFDF}\textbf{78.00} & \cellcolor[HTML]{DEDFDF}\textbf{62.95} & \cellcolor[HTML]{DEDFDF}\textbf{70.48} \\
     & \cellcolor[HTML]{DEDFDF}mChartQA$_{\text{Intern-LM2}}$  & \cellcolor[HTML]{DEDFDF}\textbf{96.06} & \cellcolor[HTML]{DEDFDF}\textbf{96.30} & \cellcolor[HTML]{DEDFDF}\textbf{96.18} & \cellcolor[HTML]{DEDFDF}\textbf{78.25} & \cellcolor[HTML]{DEDFDF}\textbf{74.79} & \cellcolor[HTML]{DEDFDF}\textbf{76.52} \\
    \bottomrule
    \end{tabular}}
 \label{tab:figureqa_plotqa_results}
\end{table}

In our analysis, we benchmark the mChartQA model against a diverse set of baselines, encompassing both Few-Shot Learning Models and Fully-Supervised Models. These baselines represent the state-of-the-art performance on the datasets in question, providing a rigorous context for evaluating the effectiveness of our model. The comparison includes models such as GPT3 and GPT4 for Few-Shot learning scenarios, specialized chart interpretation models like Pix2Struct and MATCHA, and LLM-Based Approaches such as PaLI-X-OCR. This diverse set of baselines ensures a comprehensive assessment of mChartQA's performance across different learning paradigms and task complexities.

\textbf{Performance on the ChartQA Dataset:}
As shown in Table \ref{tab:extended_chartqa}, while mChartQA demonstrates superior performance in several aspects, particularly in handling machine-generated questions with its advanced understanding of chart elements, it faces stiff competition from models like DOMINO in the fully supervised setting. The effectiveness of mChartQA is particularly notable in scenarios requiring deep semantic understanding and contextual interpretation, where it outperforms traditional models. However, in tasks where extensive pre-training on chart-specific data is beneficial, models like MATCHA and Pix2Struct show competitive advantages. This suggests that while mChartQA excels in leveraging contextual cues and integrating multimodal information, there is room for improvement in direct chart data interpretation. Future improvements will focus in this direction.

\textbf{Analysis on FigureQA and PlotQA Datasets:}
As shown in Table \ref{tab:figureqa_plotqa_results}, mChartQA's performance on the FigureQA and PlotQA datasets is robust, showcasing its generalizability across different chart types and question formats. It achieves remarkable accuracy, especially in the Intern-LM2 version, surpassing many baselines in both Few-Shot and Fully-Supervised categories. However, the model does not always lead, with certain baselines like ChartReader showing superior performance in specific tasks, indicating the potential benefits of integrating more targeted OCR and visual feature extraction techniques. This underscores the need for ongoing refinement in the model's approach to visual data processing and interpretation.

\begin{table*}[t]
\centering
\caption{Comparative Results in Complex Scenarios}
{\renewcommand\baselinestretch{1.25}\selectfont
\resizebox{1.0\textwidth}{!}{
\Huge
\begin{tabular}{l|ccc|cc|cc|c|c|c}
\toprule
\multirow{2}[1]{*}{\textbf{Model}}  &\multicolumn{3}{c|}{\textbf{ChartQA-Test}}&\multicolumn{2}{c|}{\textbf{PoltQA-Test V1}} &\multicolumn{2}{c|}{\textbf{PoltQA-Test V2}}
&\multicolumn{1}{c|}{\textbf{FigureQA-Val1}} &\multicolumn{1}{c|}{\textbf{FigureQA-Val2}} &\multicolumn{1}{c}{\textbf{Overall}}\\
& Color & Structure & Textless & Structure & Textless& Structure & Textless & Textless & Textless & avg \\
\midrule
Pix2Struct ~\citep{pix2struct}&26.90 & 28.05 & 34.45 & 25.47 & 23.43 & 19.03 &  6.80  & 52.02 & 52.00	& 29.79\\	
MATCHA~\citep{matcha} &29.55 & 32.21 & 54.55 & 22.46 & 23.52 & 19.03 & \underline{14.65}  & 50.90 & 50.71	&33.06\\	
\hline
BLIP-2 ~\citep{blip2} &  7.20 &  9.87 & 15.31 & 18.53 & 18.80 & 19.35 &  3.78 &  51.03 & 48.80	&21.41\\	
InstructBLIP  ~\citep{instructblip}& 4.55 &  9.35 & 15.79 & 32.60 & 23.86 & 25.16 &  6.50 &  43.25 & 43.58	&22.74\\	
UniChart ~\citep{unichart}&34.09 & 36.36 & \underline{62.20} & 31.68 & 27.30 & 27.74 & 12.24 &  51.34 & 51.2	&37.13\\	
Qwen-VL-Chat ~\citep{qwenvl} &\underline{37.12} & \underline{37.92} & 40.67 & \underline{48.14} & \underline{38.71} & \underline{46.13} & 13.90 &  \underline{55.01} & \underline{55.69}	&\underline{41.48}\\		
\hline
\rowcolor[HTML]{DEDFDF}  
mChartQA$_{\text{Qwen}}$  & \textbf{43.57} ($\blacktriangle$ 6.45) & \textbf{52.21} ($\blacktriangle$ 14.29) & \textbf{64.11} ($\blacktriangle$ 1.91) & \textbf{61.18} ($\blacktriangle$ 13.04) & \textbf{66.67} ($\blacktriangle$ 27.96) & \textbf{60.32} ($\blacktriangle$ 14.19)& \textbf{62.54} ($\blacktriangle$ 47.89)& \textbf{74.92} ($\blacktriangle$ 19.91)& \textbf{72.96} ($\blacktriangle$ 17.27)& \textbf{62.05} ($\blacktriangle$ 20.57)\\
\rowcolor[HTML]{DEDFDF} 	
mChartQA$_{\text{Intern-LM2}}$  & \textbf{62.50} ($\blacktriangle$ 25.38) & \textbf{62.08} ($\blacktriangle$ 24.16) & \textbf{56.94} ($\blacktriangle$ -2.56) & \textbf{89.96} ($\blacktriangle$ 41.82) & \textbf{75.71} ($\blacktriangle$ 37.00) & \textbf{87.10} ($\blacktriangle$ 40.97)& \textbf{46.22} ($\blacktriangle$ 31.57)& \textbf{95.82} ($\blacktriangle$ 40.81)& \textbf{95.70} ($\blacktriangle$ 40.01)& \textbf{74.67} ($\blacktriangle$ 33.19)\\
\bottomrule
\end{tabular}} \par}
\label{tab:chartqa_mul}
\end{table*}

\textbf{Comparative Analysis in Complex Scenarios:}
As shown in Table \ref{tab:chartqa_mul}, the comparative results in complex scenarios such as color, structure, and textless tasks showcase the capabilities of the mChartQA model. The mChartQA model, especially the Intern-LM2 version, demonstrates a remarkable ability to outperform existing models across these challenging scenarios. Notably, the model achieves significant advancements in the textless category across different datasets, which underscores its proficiency in interpreting charts that lack explicit textual annotations.
Despite these strengths, the analysis also uncovers areas where mChartQA could be further enhanced. For instance, while the model shows exceptional performance in textless scenarios, its performance in color and structure-related tasks, though superior, suggests there is still room for improvement. This is particularly evident when compared to the highest scores achieved by other models in specific categories, indicating that mChartQA's approach to processing and understanding visual elements such as color and structural components can be refined.
The comparative performance also highlights the potential for mChartQA to benefit from more targeted improvements in its handling of complex chart elements. For example, the slight underperformance in certain scenarios compared to the very best model outcomes suggests that integrating more sophisticated visual processing techniques or enhancing the model's training on datasets with a wider variety of chart types could yield further improvements.

\textbf{Overall Performance and Future Directions:}
In summary, mChartQA establishes a new standard in multimodal chart question-answering, showcasing strong performance in complex scenarios such as those involving color, structure, and charts lacking textual descriptions. Despite its notable achievements, the analysis identifies potential areas for further refinement, especially in enhancing the model's interpretative accuracy across a broader range of chart types and scenarios. Future efforts will be directed towards deepening the model's comprehension of complex visual elements and improving its interpretative capabilities, with the goal of extending mChartQA's leadership in the field of chart question-answering.

\subsection{Ablation Study}
In this ablation study, we explored the impact of incorporating Deplot in different stages of training and testing in the mChartQA$_{\text{Qwen}}$ model. We analyzed the performance variations by selectively applying Deplot, as shown in Table \ref{tab:chartqa_comparison} (see rows 1-3 in Table \ref{tab:chartqa_comparison}).

The mChartQA$_{\text{Qwen}}$ model with Deplot in both training and testing phases (row 1) demonstrates the highest performance, underscoring Deplot's significance in enhancing chart comprehension. The absence of Deplot in either phase (rows 2 and 3) leads to a noticeable decline in performance, particularly in complex scenarios like textless charts. This confirms our hypothesis about Deplot's crucial role in the model's learning process and its effectiveness in handling multimodal data and complex chart structures.

\begin{table*}[ht]
\centering
\caption{Ablation Study Results on Multimodal Chart Datasets}
{\renewcommand\baselinestretch{1.25}\selectfont
\resizebox{1.0\textwidth}{!}{
\Huge
\begin{tabular}{l|cccccc|cccc|cccc|cc|cc|cc}
\toprule
\multirow{2}[1]{*}{\textbf{mChartQA$_{\text{Qwen}}$}}  & \multicolumn{6}{c|}{\textbf{ChartQA-Test}}&\multicolumn{4}{c|}{\textbf{PoltQA-Test V1}} &\multicolumn{4}{c|}{\textbf{PoltQA-Test V2}}
&\multicolumn{2}{c|}{\textbf{FigureQA-Val1}} &\multicolumn{2}{c|}{\textbf{FigureQA-Val2}} &\multicolumn{2}{c}{\textbf{Overall}} \\
& Color & $\bigtriangleup$ & Structure & $\bigtriangleup$ & Textless & $\bigtriangleup$ & Structure & $\bigtriangleup$  & Textless & $\bigtriangleup$ & Structure & $\bigtriangleup$  & Textless & $\bigtriangleup$  & Textless & $\bigtriangleup$  & Textless & $\bigtriangleup$ & avg &  $\bigtriangleup$ \\
\midrule
Base: w/ Deplot; w/ Deplot & 43.57  & -  & \textbf{52.21} & - & \textbf{64.11} &  -  & \textbf{61.18} & - &  \textbf{66.67} & - & \textbf{60.32} & - & \textbf{62.54} & - & \textbf{74.92} & - & \textbf{72.96} & - & \textbf{62.05} &-\\	
\hline
w/o  Deplot; w/o  Deplot & 34.97  & -8.6 & 37.05 & -15.16 & 43.60 & -20.51 & 36.20 & -24.98 & 39.79 & -26.88 & 48.92 & -11.4 & 22.66 & -39.88 & 55.49 & -19.43 & 57.12	& -15.84 & 41.76 & -20.29\\
w/o  Deplot;  w/  Deplot &35.23 & -8.34& 42.08 & -10.13 & 49.76 & -14.35 & 45.54& -15.64 & 46.70& -19.97 & 45.16 & -15.16 & 29.00 & -33.54 & 56.07& -18.85& 55.99	& -16.97 & 45.06 & -16.99\\
-Qformer + MLP & \textbf{47.35} & +3.78 & 51.43 & -0.78 & 63.16 & -0.95 & 42.96 & -18.22 & 53.65 & -13.02 & 40.65 & -19.67 & 45.62 & -16.92 & 66.58 & -8.34 & 64.78	& -8.18 & 52.91 & -9.14\\
-ViT448 + Vit384 &46.21&+2.64 & 51.17 & -1.04 & 63.64 & -0.47 & 53.10& -8.08 & 57.25 & -9.42 & 51.61 & -8.71 & 49.85 & -12.69 & 68.45 & -6.44 & 64.97	& -7.99 & 56.25 & -5.8\\
\bottomrule
\end{tabular}} \par}
\label{tab:chartqa_comparison}
\end{table*}

\subsection{Further Analysis}
\textbf{Effect of Connector Replacement (see rows 4 in Table \ref{tab:chartqa_comparison}):} Replacing our cross-attention connector with an MLP-based approach (-Qformer + MLP) resulted in varied performance across different tasks. While the MLP connector showed some improvements in specific scenarios, such as color charts in ChartQA, our original connector generally outperformed the MLP in most tasks, especially in complex reasoning scenarios. This highlights the effectiveness of our cross-attention mechanism in integrating multimodal information.

\textbf{Effect of Visual Encoder Variation (see rows 5 in Table \ref{tab:chartqa_comparison}):} Experimenting with different visual encoders, we replaced our ViT-448 encoder with ViT-384 (-ViT448 + Vit384). The results indicate that while ViT-384 performs competitively in certain tasks, our ViT-448 encoder generally achieves superior results, particularly in handling complex chart structures and textless scenarios. This suggests the importance of a specialized visual encoder tailored for multimodal chart question-answering.
\subsection{Case Study}
\begin{figure}[ht]
    \centering
    \includegraphics[width=0.6\linewidth]{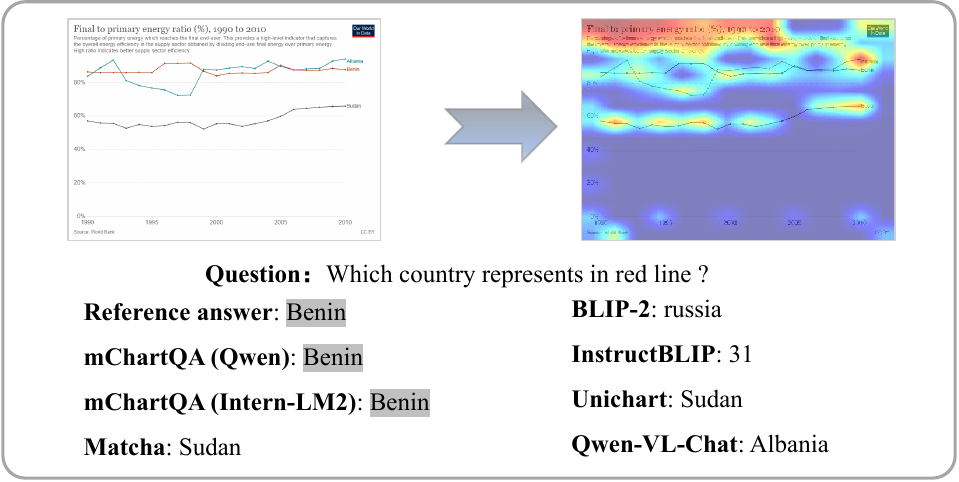}
    \caption{Case study example.}
    \label{fig:case_study}
\end{figure}
In this case study, we present an example that highlights the superior performance of our mChartQA model, especially in complex scenarios involving overlapping visual elements. Figure \ref{fig:case_study} illustrates this case,

This example demonstrates mChartQA (Base)'s ability to accurately interpret complex charts with significant visual element overlap. mChartQA is the only model among its comparisons to correctly identify the chart's details. We speculate that this success is mainly due to the multimodal architecture, where the Vision Encoder provides precise information to the language encoder. The heatmap visualization confirms our speculation. In contrast, other models exhibit severe hallucinations, either misinterpreting or providing arbitrary answers.

\section{Error Analysis}

\begin{figure*}[htbp]
\centering
\includegraphics[width=\linewidth]{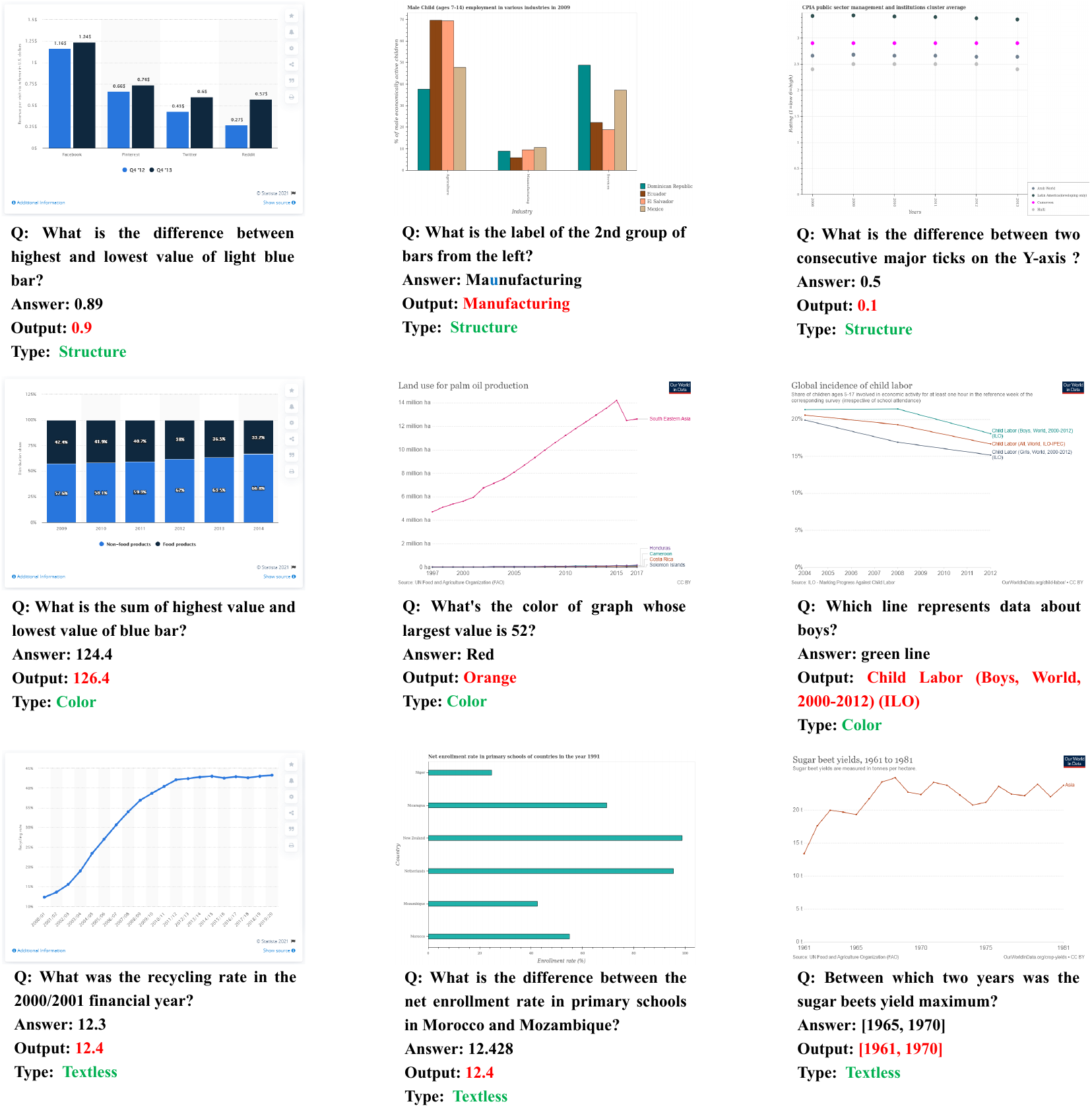}
\caption{Examples of errors in structure, color, and textless chart questions.}
\label{fig:error_analysis_examples}
\end{figure*}

We delve into a detailed error analysis of our model, categorizing errors into three main types: structure, color, and textless chart questions. We identified prevalent error patterns within each category, as demonstrated in Figure \ref{fig:error_analysis_examples}.

\textbf{Structure-Related Errors:} In the first row of Figure \ref{fig:error_analysis_examples}, we present errors encountered in structure-related chart questions. For example, the first error was due to rounding off the answer, resulting in a slight deviation from the correct value. The second error involved a language model hallucination, leading to a misspelling. The third error was attributed to inadequate structural recognition, affecting the accuracy of the response.

\textbf{Color-Related Errors:} The second row in Figure \ref{fig:error_analysis_examples} shows errors in color-related chart questions. The first error occurred in multi-step calculations, where a mistake at any stage could lead to an incorrect outcome. The second error was due to limited color recognition accuracy, leading to ambiguity in identifying the correct color. The third error highlighted the need for precise question interpretation, as the model's response, although technically correct, did not align with the expected answer format.

\textbf{Textless Chart Errors:} The third row of Figure \ref{fig:error_analysis_examples} illustrates errors in textless chart questions. These included slight deviations in numerical estimations, precision mismatches between the model's output and the standard answer, and confusion in the model's understanding of the question scope.

This error analysis sheds light on potential areas for refinement in multimodal chart question-answering models. The issues observed in numerical precision, language model accuracy, and recognition in structural and color aspects will be the focus of our continued optimization efforts in the future.

\section{Conclusion}
In this study, our mChartQA model, designed to address the complex challenges of multimodal chart question-answering, particularly excelled in scenarios involving complex color patterns, structural complexities, and interpreting charts with implicit numerical data. By leveraging a two-stage training strategy, mChartQA achieved state-of-the-art performance in diverse chart question-answering tasks, demonstrating the effectiveness of our approach. Looking ahead, we plan to further refine the visual encoder and connector components within mChartQA to enhance its multimodal chart question-answering capabilities.

\bibliographystyle{cas-model2-names}

\bibliography{cas-refs}

\end{document}